\documentclass[twoside,11pt]{article}

\usepackage{dnd}
\usepackage{times}
\usepackage{amsmath, amsfonts}
\usepackage{graphicx}
\usepackage{lscape}

\dndheading{7(3)}{2016}{34 --46}{Julien Perez}{10.5087/dad.2016.304}

\ShortHeadings{Spectral State Tracker}{{Perez}, Julien}
\title{Spectral decomposition method of dialog state tracking via collective matrix factorization}

\author{ \\
Xerox Research Centre Europe, France \\
{\tt Julien.Perez@xerox.com}\\}

\author{\name Julien Perez \email julien.perez@xrce.xerox.com \\
       \addr Xerox Research Center Europe\\
       Grenoble, 38000, France
	}

\editor{Jason D. Williams, Antoine Raux, and Matthew Henderson}
\submitted{04/15}{02/16}{04/16}

\DeclareMathOperator*{\argmin}{arg\,min}

\firstpageno{34}
\begin{document}
\maketitle


\begin{abstract}

The task of dialog management is commonly decomposed into two sequential subtasks: dialog state tracking and dialog policy learning. In an end-to-end dialog system, the aim of dialog state tracking is to accurately estimate the true dialog state from noisy observations produced by the speech recognition and the natural language understanding modules. The state tracking task is primarily meant to support a dialog policy. From a probabilistic perspective, this is achieved by maintaining a posterior distribution over hidden dialog states composed of a set of context dependent variables. Once a dialog policy is learned, it strives to select an optimal dialog act given the estimated dialog state and a defined reward function. This paper introduces a novel method of dialog state tracking based on a bilinear algebric decomposition model that provides an efficient inference schema through collective matrix factorization. We evaluate the proposed approach on the second Dialog State Tracking Challenge (DSTC-2) dataset and we show that the proposed tracker gives encouraging results compared to the state-of-the-art trackers that participated in this standard benchmark. Finally, we show that the prediction schema is computationally efficient in comparison to the previous approaches.

\end{abstract}


\section{Introduction}

The field of autonomous dialog systems is rapidly growing with the spread of smart mobile devices but it still faces many challenges to become the primary user interface for natural interaction through conversations. Indeed, when dialogs are conducted in noisy environments or when utterances themselves are noisy, correctly recognizing and understanding user utterances presents a real challenge. In the context of call-centers, efficient automation has the potential to boost productivity through increasing the probability of a call's success while reducing the overall cost of handling the call. One of the core components of a state-of-the-art dialog system is a dialog state tracker. Its purpose is to monitor the progress of a dialog and provide a compact representation of past user inputs and system outputs represented as a dialog state. The dialog state encapsulates the information needed to successfully finish the dialog, such as users' goals or requests. Indeed, the term ``dialog state'' loosely denotes an encapsulation of user needs at any point in a dialog. Obviously, the precise definition of the state depends on the associated dialog task. An effective dialog system must include a tracking mechanism which is able to accurately accumulate evidence over the sequence of turns of a dialog, and it must adjust the dialog state according to its observations. In that sense, it is an essential componant of a dialog systems. However, actual user utterances and corresponding intentions are not directly observable due to errors from Automatic Speech Recognition (ASR) and Natural Language Understanding (NLU), making it difficult to infer the true dialog state at any time of a dialog. A common method of modeling a dialog state is through the use of a slot-filling schema, as reviewed in \cite{WilliamsY07}. In slot-filling, the state is composed of a predefined set of variables with a predefined domain of expression for each of them. The goal of the dialog system is to efficiently instantiate each of these variables thereby performing an associated task and satisfying the corresponding intent of the user.

Various approaches have been proposed to define dialog state trackers. The traditional methods used in most commercial implementations use hand-crafted rules that typically rely on the most likely result from an NLU module as described in \cite{YehDJRRPLTBM14}. However, these rule-based systems are prone to frequent errors as the most likely result is not always the correct one. Moreover, these systems often force the human customer to respond using simple keywords and to explicitly confirm everything they say, creating an experience that diverges considerably from the natural conversational interaction one might hope to achieve as recalled in \cite{Williams14}. More recent methods employ statistical approaches to estimate the posterior distribution over the dialog states allowing them to represent the uncertainty of the results of the NLU module. Statistical dialog state trackers are commonly categorized into one of two approaches according to how the posterior probability distribution over the state calculation is defined. In the first type, the generative approach uses a generative model of the dialog dynamic that describes how the sequence of utterances are generated by using the hidden dialog state and using Bayes' rule to calculate the posterior distribution of the state. It has been a popular approach for statistical dialog state tracking, since it naturally fits into the Partially Observable Markov Decision Process (POMDP) models as described in \cite{YoungGTW13}, which is an integrated model for dialog state tracking and dialog strategy optimization. Using this generic formalism of sequential decision processes, the task of dialog state tracking is to calculate the posterior distribution over an hidden state given an history of observations. In the second type, the discriminative approach models the posterior distribution directly through a closed algebraic formulation as a loss minimization problem. Statistical dialog systems, in maintaining a distribution over multiple hypotheses of the true dialog state, are able to behave robustly even in the face of noisy conditions and ambiguity. In this paper, a statistical type of approach of state tracking is proposed by leveraging the recent progress of spectral decomposition methods formalized as bilinear algebraic decomposition and associated inference procedures. The proposed model estimates each state transition with respect to a set of observations and is able to compute the state transition through an inference procedure with a linear complexity with respect to the number of variables and observations.

{\bf Roadmap: }This paper is structured as follows, Section \ref{sec:pb} formally defines transactional dialogs and describes the associated problem of statistical dialog state tracking with both the generative and discriminative approaches. Section \ref{sec:prop} depicts the proposed decompositional model for coupled and temporal hidden variable models and the associated inference procedure based on Collective Matrix Factorization (CMF). Finally, Section \ref{sec:xp} illustrates the approach with experimental results obtained using a state of the art benchmark for dialog state tracking.


\section{Transactional dialog state tracking}
\label{sec:pb}

\begin{figure}[ht!]
\centering
\includegraphics[width=0.35\textwidth]{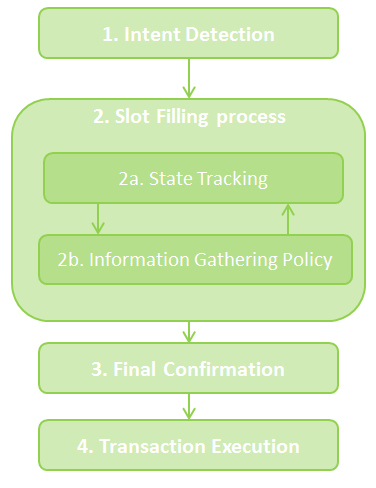}
\caption{Prototypical transactional dialog management process, also called slot-filling dialog management}
\label{fig:slotfilling}
\end{figure}

The dialog state tracking task we consider in this paper is formalized as follows: at each turn of a task-oriented dialog between a dialog system and a user, the dialog system chooses a dialog act $d$ to express and the user answers with an utterance $u$. The dialog state at each turn of a given dialog is defined as a distribution over a set of predefined variables, which define the structure of the state as mentioned in \cite{Williams05}. This classic state structure is commonly called {\it slot filling} and the associated dialogs are commonly referred to as {\it transactional}. Indeed, in this context, the state tracking task consists of estimating the value of a set of predefined variables in order to perform a procedure or transaction which is, in fact, the purpose of the dialog. Typically, the NLU module processes the user utterance and generates an N-best list $o = \{<d_1, f_1>, \ldots , <d_n, f_n>\}$, where $d_i$ is the hypothesized user dialog act and $f_i$ is its confidence score. In the simplest case where no ASR and NLU modules are employed, as in a text based dialog system as proposed in \cite{Henderson13a} the utterance is taken as the observation using a so-called bag of words representation. If an NLU module is available, standardized dialog act schemas can be considered as observations as in \cite{bunt10}. Furthermore, if prosodic information is available by the ASR component of the dialog system as in \cite{MiloneR03}, it can also be considered as part of the observation definition. A statistical dialog state tracker maintains, at each discrete time step $t$, the probability distribution over states, $b(s_t)$, which is the system's {\it belief} over the state. The general process of slot-filling, transactional dialog management is summarized in Figure \ref{fig:slotfilling}. First, {\it intent detection} is typically an NLU problem consisting of identifying the task the user wants the system to accomplish. This first step determines the set of variables to instantiate during the second step, which is the slot-filling process. This type of dialog management assumes that a set of variables are required for each predefined intention. The slot filling process is a classic task of dialog management and is composed of the cyclic tasks of {\it information gathering} and integration, in other words -- {\it dialog state tracking}. Finally, once all the variables have been correctly instantiated, a common practice in dialog systems is to perform a last general confirmation of the task desired by the user before finally executing the requested task. As an example used as illutration of the proposed method in this paper, in the case of the DSTC-2 challenge, presented in \cite{hen14}, the context was taken from the restaurant information domain and the considered variables to instanciate as part of the state are \{Area (5 possible values) ; FOOD (91 possible values) ;  Name (113 possible values) ;  Pricerange (3 possible values)\}. In such framework, the purpose is to estimate as early as possible in the course of a given dialog the correct instantiation of each variable. In the following, we will assume the state is represented as a concatenation of zero-one encoding of the values for each variable defining the state. Furthermore, in the context of this paper, only the bag of words has been considered as an observation at a given turn but dialog acts or detected named entity provided by an SLU module could have also been incorporated as evidence.

Two statistical approaches have been considered for maintaining the distribution over a state given sequential NLU output. First, the discriminative approach aims to model the posterior probability distribution of the state at time $t+1$ with regard to state at time $t$ and observations $z_{1:t}$. Second, the generative approach attempts to model the transition probability and the observation probability in order to exploit possible interdependencies between hidden variables that comprise the dialog state.

\subsection{Generative Dialog State Tracking}

A generative approach to dialog state tracking computes the belief over the state using Bayes' rule, using the belief from the last turn $b(s_{t-1})$ as a prior and the likelihood given the user utterance hypotheses $p(z_t|s_t)$, with $z_t$ the observation gathered at time $t$. In the prior work \cite{Williams05}, the likelihood is factored and some independence assumptions are made:

\begin{equation}
b_t \propto \sum_{s_{t-1},z_t} p(s_t|z_t, d_{t-1}, s_{t-1}) p(z_t|s_t)  b(s_{t-1})
\end{equation}

\begin{figure}
\centering
\includegraphics[width=0.4\textwidth]{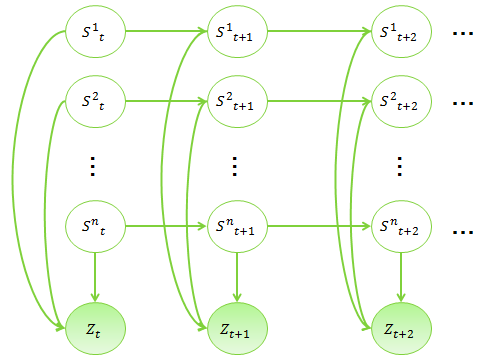}
\caption{Generative Dialog State Tracking using a factorial HMM}
\label{fig:fhmm}
\end{figure}

Figure \ref{fig:fhmm} depicts a typical generative model of a dialog state tracking process using a factorial hidden Markov model proposed by \cite{gj97}. The shaded variables are the observed dialog turns and each unshaded variable represents a single variable describing the task dependent variables. In this family of approaches, scalability is considered as one of the main issues. One way to reduce the amount of computation is to group the states into partitions, as proposed in the Hidden Information State (HIS) model of \cite{GasicY11}. Other approaches to cope with the scalability problem in dialog state tracking is to adopt a factored dynamic Bayesian network by making conditional independence assumptions among dialog state components, and then using approximate inference algorithms such as loopy belief propagation as proposed in \cite{ThomsonY10} or a blocked Gibbs sampling as in \cite{RauxM11}. To cope with such limitations, discriminative methods of state tracking presented in the next part of this section aim at directly model the posterior distribution of the tracked state using a choosen parametric form.

\subsection{Discriminative Dialog State Tracking}

The discriminative approach of dialog state tracking computes the belief over a state via a trained parametric model that directly represents the belief $b(s_{t+1}) = p(s_{s+1} | s_t, z_t)$. Maximum Entropy has been widely used in the discriminative approach as described in \cite{MetallinouBW13}. It formulates the belief as follows:

\begin{equation}
b(s) = P(s|x) = \eta.e^{w^T\phi(x,s)}
\end{equation}

where $\eta$ is the normalizing constant, $x = (d^u_1, d^m_1, s_1, \dots, d^u_t, d^m_t, s_t)$ is the history of user dialog acts, $d^u_i, i \in \{1,\ldots,t\}$, the system dialog acts, $d^m_i, i \in \{1,\ldots,t\}$, and the sequence of states leading to the current dialog turn at time $t$. Then, $\phi(.)$ is a vector of feature functions on $x$ and $s$, and finally, $w$ is the set of model parameters to be learned from annotated dialog data. According to the formulation, the posterior computation has to be carried out for all possible state realizations in order to obtain the normalizing constant $\eta$. This is not feasible for real dialog domains, which can have a large number of variables and possible variable instantiations. So, it is vital to the discriminative approach to reduce the size of the state space. For example, \cite{MetallinouBW13} proposes to restrict the set of possible state variables to those that appeared in NLU results. More recently, \cite{LeeJKRL13} assumes conditional independence between dialog state variables to address scalability issues and uses a conditional random field to track each variable separately. Finally, deep neural models, performing on a sliding window of features extracted from previous user turns, have also been proposed in \cite{Henderson14}. Of the current literature, this family of approaches have proven to be the most efficient for publicly available state tracking datasets. In the next section, we present a decompositional approach of dialog state tracking that aims at reconciling the two main approaches of the state of the art while leveraging on the current advances of low-rank bilinear decomposition models, as recalled in \cite{MaBRC14}, that seems particularly adapted to the sparse nature of dialog state tracking tasks.


\section{Spectral decomposition model for state tracking in slot-filling dialogs}
\label{sec:prop}

\begin{figure}[ht!]
\centering
\includegraphics[width=0.6\textwidth]{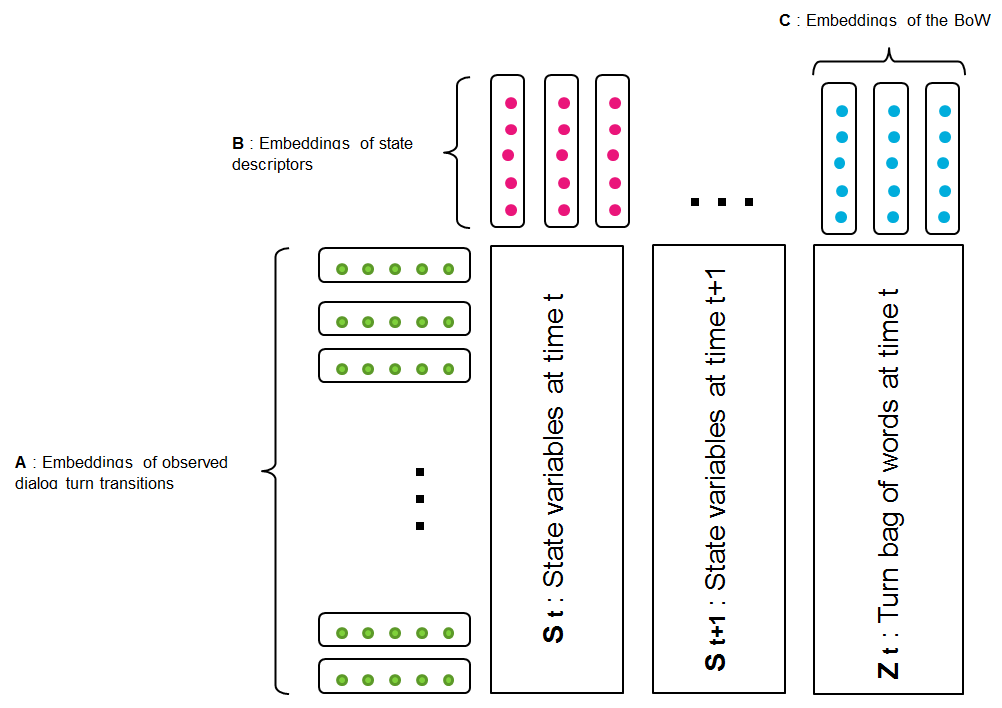}
\caption{Spectral State Tracking, Collective Matrix Factorization model as inference procedure}
\label{fig:modelimpl}
\end{figure}

In this section, the proposed model is presented and the learning and prediction procedures are detailed. The general idea consists in the decomposition of a matrix $M$, composed of a set of turn's transition as rows and sparse encoding of the corresponding feature variables as columns. More precisely, a row of $M$ is composed with the concatenation of the sparse representation of (1) $s_{t}$, a state at time $t$ (2) $s_{t+1}$, a state at time $t+1$ (3) $z_t$, a set of feature representating the observation. In the considered context, the bag of words composing the current turn is chosen as the observation. The parameter learning procedure is formalized as a matrix decomposition task solved through Alternating Least Square Ridge regression. The ridge regression task allows for an asymmetric penalization of the targeted variables of the state tracking task to perform. Figure \ref{fig:modelimpl} illustrates the collective matrix factorization task that constitutes the learning procedure of the state tracking model. The model introduces the component of the decomposed matrix to the form of latent variables $\{A, B, C\}$, also called embeddings. In the next section, the learning procedure from dialog state transition data and the proper tracking algorithm are described. In other terms, each row of the matrix corresponds to the concatenation of a "one-hot" representation of a state description at time $t$ and a dialog turn at time $t$ and each column of the overall matrix $M$ corresponds to a consider feature respectively of the state and dialog turn. Such type of modelization of the state tracking problem presents several advantages. First, the model is particularly flexible, the definition of the state and observation spaces are independent of the learning and prediction models and can be adapted to the context of tracking. Second, a bias by data can be applied in order to condition the transition model w.r.t separated matrices to decompose jointly as often proposed in multi-task learning as described in \cite{car96} and collective matrix factorization as detailed in \cite{BokdeGM15a}. Finally, the decomposition method is fast and parallelizable because it mainly leverages on core methods of linear algebra. From our knowledge, this proposition is the first attend to formalize and solve the state tracking task using a matrix decomposition approach.

\subsection{Learning method}

 For the sake of simplicity, the $\{B,C\}$ matrices are concatenated to $E$, and $M$ is the concatenation of the matrices $\{S_t,S_{t+1},Z_t\}$ depicted in Figure \ref{fig:modelimpl}. Equation \ref{eq:eq1} defines the optimization task, i.e. the loss function, associated with the learning problem of latent variable search $\{A,E\}$.

\begin{small}
\begin{equation}
 \min_{A,E} ||  (M - AE ) W||_2^2 + \lambda_a ||A||_2^2 + \lambda_b ||E||_2^2
 \enspace,
\label{eq:eq1}
\end{equation}
\end{small}

where $\{\lambda_a, \lambda_b\} \in \mathbb{R}^2$ are regularization hyper-parameters and $W$ is a diagonal matrix that increases the weight of the state variables, $s_{t+1}$ in order bias the resulting parameters $\{A,E\}$ toward better predictive accuracy on these specific variables. This type of weighting approach has been shown to be as efficient in comparable generative discriminative trade-off tasks as mentioned in \cite{Bishop06} and \cite{Bishop07}. An \emph{Alternating Least Squares} method that is a sequence of two convex optimization problems is used in order to perform the minimization task. First, for known $E$, compute:

\begin{small}
\begin{equation}
A^* = \argmin_{A} || (M - AE ) W ||_2^2 + \lambda_a ||A||_2^2
 \enspace,
\end{equation}
\end{small}

then for a given $A$, 
\begin{small}
\begin{equation}
E^* = \argmin_{E} || (M - AE) W ||_2^2 + \lambda_b ||E||_2^2 
\end{equation}
\end{small}

By iteratively solving these two optimization problems, we obtain the following fixed-point regularized and weighted alternating least square algorithms where $t$ correspond to the current step of the overall iterative process:

\begin{small}
\begin{equation}
\label{eq:a}
A_{t+1} \leftarrow (E_{t}^TWE_{t} + \lambda_a\mathbb{I})^{-1}E_{t}^TWM
\end{equation}
\end{small}

\begin{small}
\begin{equation}
\label{eq:b}
E_{t+1} \leftarrow (A_{t}^TA_{t} + \lambda_b\mathbb{I})^{-1}A_{t}^TM
\end{equation}
\end{small}

As presented in Equation~\ref{eq:a}, the $W$ matrix is only involved for the updating of $A$ because only the subset of the columns of $E$, representing the features of the state to predict, are weighted differently in order to  increase the importancd of the corresponding columns in the loss function. For the optimization of the latent representation composing $E$, presented in Equation~\ref{eq:b}, each call session's embeddings stored in $A$ hold the same weight, so in this second step of the algorithm, $W$ is actually an identity matrix and so does not appear.

\subsection{Prediction method}

The prediction process consists of (1) computing the embedding of a current transition by solving the corresponding least square problem based on the two variables $\{s_t,z_t\}$ that correspond to our current knowledge of the state at time $t$ and the set of observations extracted from the last turn that is composed with the system and user utterances, (2) estimating the missing values of interest, i.e. the likelihood of each value of each variable that constitutes the state at time $(t+1)$, $s_{t+1}$, by computing the cross-product between the transition embedding calculated in (1) and the corresponding column embeddings of $E$, and of the value of each variable of $s_{t+1}$. More precisely, we write this decomposition as

\begin{small}
\begin{equation}
\label{eq:c}
M = A.E^T
\end{equation}
\end{small}

where $M$ is the matrix of data to decompose and $.$ the matrix-matrix product operator. As in the previous section, $A$ has a row for each transition embedding, and $E$ has a column for each variable-value embedding in the form of a zero-one encoding. When a new row of observations $m_i$ for a new set of variables state $s_i$ and observations $z_i$ and $E$ is fixed, the purpose of the prediction task is to find the row $a_i$ of $A$ such that:

\begin{small}
\begin{equation}
\label{eq:d}
a_i.E^T \approx m^T_i
\end{equation}
\end{small}

Even if it is generally difficult to require these to be equal, we can require that these last elements have the same projection into the latent space:
\begin{small}
\begin{equation}
\label{eq:e}
a_i^T.E^T.E = m_i^T.E
\end{equation}
\end{small}

Then, the classic closed form solution of a linear regression task can be derived:
\begin{small}
\begin{align}
\label{eq:f}
a_i^T = m_i^T.E.(E^T.E)^{-1} \\
a_i = (E^T.E)^{-1}.E^T.m_i
\end{align}
\end{small}

In fact, Equation \ref{eq:f} is the optimal value of the embedding of the transition $m_i$, assuming a quadratic loss is used. Otherwise it is an approximation, in the case of a matrix decomposition of $M$ using a logistic loss for example. Note that, in equation \ref{eq:f}, $
(E^T.E)^{-1}$ requires a matrix inversion, but for a low dimensional matrix (the size of the latent space). Several advantages can be identified in this approach. First, at learning time, alternative ridge regression is computationally efficient because a closed form solution exists at each step of the optimization process employed to infer the parameters, i.e the low rank matrices, of the model. Second, at decision time, the state tracking procedure consists of (1) computing the embedding $a$ of the current transition using the current state estimation $s_t$ and the current observation set $z_t$ and (2) computing the distribution over the state defined as a vector-matrix product between $a$ and the latent matrix $E$.  Finally, this inference method can be partially associated to the general technique of matrix completion. But, a proper matrix completion task would have required a matrix $M$ with missing value corresponding to the exhausive list of the possible triples ${s_t, s_{t+1}, z_t}$, which is obviously intractable to represent and decompose.


\section{Experimental settings and Evaluation}
\label{sec:xp}

In a first section, the dialog domain used for the evaluation of our dialog tracker is described and the different probability models used for the domain. In a second section, we present a first set of experimental results obtained through the proposed approach and its comparison to several reported results of approaches of the state of the art. 

 \subsection{Restaurant information domain}

We used the {\it DSTC-2} dialog domain as described in \cite{williams13a} in which the user queries a database of local restaurants by interacting with a dialog system. The dataset for the restaurant information domain were originally collected using Amazon Mechanical Turk. A usual dialog proceeds as follows: first, the user specifies his personal set of constraints concerning the restaurant he looks for. Then, the system offers the name of a restaurant that satisfies the constraints. User then accepts the offer, and requests for additional information about accepted restaurant. The dialog ends when all the information requested by the user are provided. In this context, the dialog state tracker should be able to track several types of information that composes the state like the geographic area, the food type, the name and the price range slots. In this paper, we restrict ourselves to tracking these variables, but our tracker can be easily setup to track others as well if they are properly specified. The dialog state tracker updates its belief turn by turn, receiving evidence from the NLU module with the actual utterance produced by the user. In this experiment, it has been chosen to restrict the output of the NLU module to the bag of word of the user utterances in order to be comparable the most recent approaches of state tracking like proposed in  \cite{Henderson13a} that only use such information as evidence. One important interest in such approach is to dramatically simplify the process of state tracking by suppressing the NLU task. In fact, NLU is mainly formalized in current approaches as a supervised learning approach. The task of the dialog state tracker is to generate a set of possible states and their confidence scores for each slot, with the confidence score corresponding to the posterior probability of each variable state w.r.t the current estimation of the state and the current evidence. Finally, the dialog state tracker also maintains a special variable state, called {\it None}, which represents that a given variable composing the state has not been observed yet. For the rest of this section, we present experimental results of state tracking obtained in this dataset and we compare with state of the art generative and discriminative approaches.

 \subsection{Experimental results}

\begin{table}
\begin{center}
\begin{tabular}{|l|r|}
\hline
Tracker & Joint goal \\
\hline
Baseline & 0.69  \\
Focus & 0.74 \\
HWU  & 0.75  \\
HWU+ & 0.72 \\
Rule-based & 0.73 \\
MaxEnt & 0.72  \\
RNN & 0.75 \\
\hline
CMF-350    & $0.79 \pm 0.03$ \\
\hline
\end{tabular}
\caption{Accuracy of the proposed model on the DSTC-2 test-set}
\label{fig:xpres}
\end{center}
\end{table}

As a comparison to the state of the art methods, Table \ref{fig:xpres} presents accuracy results of the best Collective Matrix Factorization model, with a latent space dimension of $350$, which has been determined by cross-validation on a development set, where the value of each slot is instantiated as the most probable w.r.t the inference procedure presented in Section \ref{sec:prop}. In our experiments, the variance is estimated using standard dataset reshuffling. The same results are obtained for several state of the art methods of generative and discriminative state tracking on this dataset using the publicly available results as reported in \cite{SunCZY14}. More precisely, as provided by the state-of-the-art approaches, the accuracy scores computes $p(s^*_{t+1}|s_t,z_t)$ commonly name the joint goal. Our proposition is compared to the 4 baseline trackers provided by the DSTC organisers. They are the baseline tracker (Baseline), the focus tracker (Focus), the HWU tracker (HWU) and the HWU tracker with “original” flag set to (HWU+) respectively. Then a comparison to a maximum entropy (MaxEnt) proposed in ~\cite{lee2013} type of discriminative model and finally a deep neural network (DNN) architecture proposed in ~\cite{Sun14} as reported also in \cite{SunCZY14} is presented.


\section{Related work}

As depicted in Section \ref{sec:pb}, the litterature of the domain can mainly decomposed into three family of approaches, rule-based, generative and discriminative. In previous works on this topics, \cite{Williams07} formally used particle filters to perform inference in a Bayesian network modeling of the dialog state, \cite{Williams08} presented a generative tracker and showed how to train an observation model from transcribed data, \cite{Williams10a} grouped indistinguishable dialog states into partitions and consequently performed dialog state tracking on these partitions instead of the individual states, \cite{ThomsonY10} used a dynamic Bayesian network to represent the dialog model in an approximate form. So, most attention in the dialog state belief tracking literature has been given to generative Bayesian network models until recently as proposed in ~\cite{PaekH00} and ~\cite{ThomsonY10}. On the other hand, the successful use of discriminative models for belief tracking has recently been reported by ~\cite{Williams12} and ~\cite{Henderson13a} and was a major theme in the results of the recent edition of the Dialog State Tracking Challenge. In this paper, a latent decomposition type of approach is proposed in order to address this general problem of dialog system. Our method gives encouraging results in comparison to the state of the art dataset and also does not required complex inference at test time because, as detailed in Section \ref{sec:prop}, the tracking algorithm hold a linear complexity w.r.t the sum of realization of each considered variables defining the state to track which is what we believe is one of the main advantage of this method. Secondly collective matrix factorization paradigm also for data fusion and bias by data type of modeling as successfully performed in matrix factorization based recommender systems \cite{KorenBV09}.


\section{Conclusion}

In this paper, a methodology and algorithm for efficient state tracking in the context of slot-filling dialogs has been presented. The proposed probabilistic model and inference algorithm allows efficient handling of dialog management in the context of classic dialog schemes that constitute a large part of task-oriented dialog tasks. More precisely, such a system allows efficient tracking of hidden variables defining the user goal using any kind of available evidence, from utterance bag-of-words to the output of a Natural Language Understanding module. Our current investigation on this subject are the beneficiary of distributional word representation as proposed in \cite{MikolovYZ13} to cope with the question of unknown words and unknown slots as suggested in \cite{Henderson2014d}. In summary, the proposed approach differentiates itself by the following points from the prior art: (1) by producing a joint probability model of the hidden variable transition in a given dialog state and the observations that allow tracking the current beliefs about the user goals while explicitly considering potential interdependencies between state variables (2) by proposing the necessary computational framework, based on collective matrix factorization, to efficiently infer the distribution over the state variables in order to derive an adequate dialog policy of information seeking in this context. Finally, while transactional dialog tracking is mainly useful in the context of autonomous dialog management, the technology can also be used in dialog machine reading and knowledge extraction from human-to-human dialog corpora as proposed in the fourth edition of the Dialog State Tracking Challenge.

\bibliography{biblio}
 
\end{document}